\documentclass[11pt]{article}

\usepackage[preprint]{acl}

\usepackage{times}
\usepackage{latexsym}

\usepackage[T1]{fontenc}

\usepackage[utf8]{inputenc}

\usepackage{microtype}

\usepackage{inconsolata}

\usepackage{graphicx}

\title{EvalSense: A Framework for Domain-Specific LLM (Meta-)Evaluation}

\author{Adam Dejl \\
  Imperial College London\thanks{Work done while at NHS England.} \\
  Department of Computing \\
  \texttt{ad5518@ic.ac.uk} \\\And
  Jonathan Pearson \\
  NHS England \\
  Transformation Directorate \\
  \texttt{jonathanpearson@nhs.net} \\}

\usepackage{enumitem}
\usepackage{booktabs}
\usepackage{todonotes}
\usepackage{adjustbox}
\usepackage{subcaption}
\usepackage{listings}
\usepackage{float}
\newfloat{lstfloat}{htbp}{lop}
\floatname{lstfloat}{Listing}

\begin{document}
\maketitle
\begin{abstract}
Robust and comprehensive evaluation of large language models (LLMs) is essential for identifying effective LLM system configurations and mitigating risks associated with deploying LLMs in sensitive domains. However, traditional statistical metrics are poorly suited to open-ended generation tasks, leading to growing reliance on LLM-based evaluation methods. These methods, while often more flexible, introduce additional complexity: they depend on carefully chosen models, prompts, parameters, and evaluation strategies, making the evaluation process prone to misconfiguration and bias. In this work, we present EvalSense, a flexible, extensible framework for constructing domain-specific evaluation suites for LLMs. EvalSense provides out-of-the-box support for a broad range of model providers and evaluation strategies, and assists users in selecting and deploying suitable evaluation methods for their specific use-cases. This is achieved through two unique components: (1) an \emph{interactive guide} aiding users in evaluation method selection and (2) \emph{automated meta-evaluation tools} that assess the reliability of different evaluation approaches using perturbed data. We demonstrate the effectiveness of EvalSense in a case study involving the generation of clinical notes from unstructured doctor-patient dialogues, using a popular open dataset. All code, documentation, and assets associated with EvalSense are open-source and publicly available at \url{https://github.com/nhsengland/evalsense}.
\end{abstract}

\section{Introduction}
\label{sec:introduction}
Backed by training on unprecedentedly large quantities of data, large language models (LLMs) have radically advanced the field of machine learning and demonstrated a wide range of impressive capabilities across diverse domains \citep{bubeck-2023-sparks-gai,vanveen-2024-llms-outperform-summaries, luo-2025-llms-surpass-neuroscience,mcduff-2025-differential-diagnosis}. While these results suggest that LLMs have the potential to deliver substantial benefits, their use also entails significant risks, including hallucinations \citep{huang-2025-hallucinations-survey}, omissions of crucial information \citep{busch-2025-llm-applications-patient-care}, unintended disclosure of sensitive personal data \citep{das-2025-security-privacy-challenges}, and vulnerability to harmful instructions \citep{das-2025-security-privacy-challenges}. Rigorous evaluation of LLMs has been proposed as a key strategy for mitigating these risks and ensuring that LLM-based systems perform reliably on their assigned tasks \citep{who-2023-safe, ong-2024-llm-challenges}.

However, reliable evaluation of open-ended texts produced by LLMs remains challenging as a result of the unstructured and complex nature of these texts. Due to the inadequacy of standard statistical metrics, the community has increasingly adopted LLM-as-a-judge approaches \citep{liu-2023-g-eval, fu-2024-gptscore, kim-2024-prometheus, kim-2024-prometheus-2}, which use LLMs themselves to assess model outputs. These methods tend to be more effective at capturing content-related nuances and generally achieve higher correlations with human judgements \citep{lianmin-2024-judging-judges}. Yet, the reliability of LLMs as evaluators may vary depending on the considered task, LLM judge and the used evaluation strategy \citep{bhuvanashree-2025-evaluating-evaluator, tan-2025-judgebench, han-2025-judges-verdict}. This motivates the need to carefully choose the evaluation approach suitable for the specific domain and to rigorously \emph{meta-evaluate} its effectiveness (i.e., to evaluate the evaluator), steps that are often neglected in the existing evaluation pipelines.

\begin{figure*}[tb]
    \centering
    \includegraphics[width=0.95\textwidth]{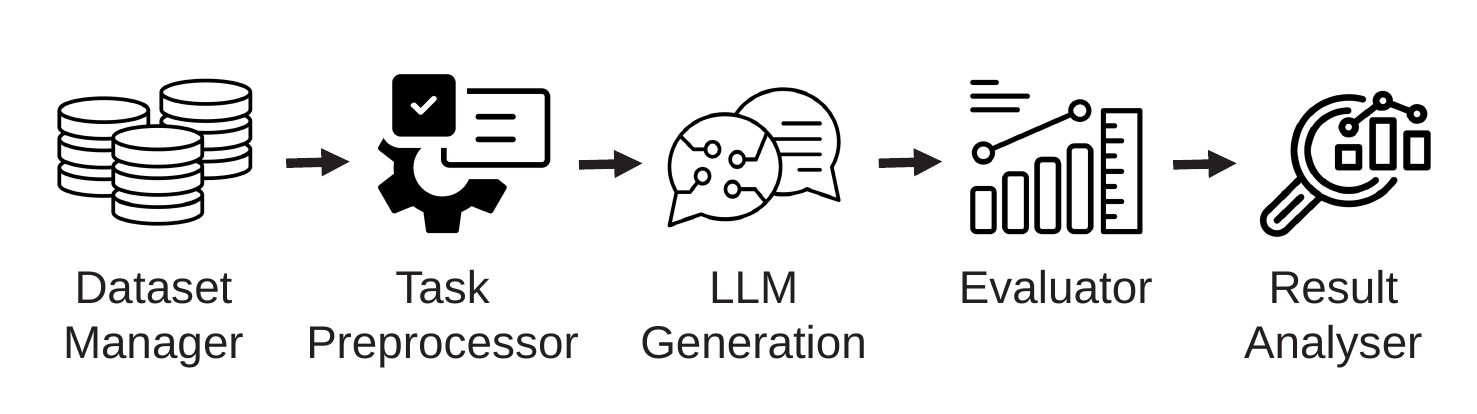}
    \caption{Overview of the LLM evaluation pipeline implemented in EvalSense\protect\footnotemark. After data loading and task-specific preprocessing, model outputs are generated and scored using different evaluators. Result analysers summarise outcomes across experiments, identify higher-level patterns, and support meta-evaluation.}
    \label{fig:pipeline}
\end{figure*}

Several open-source toolkits and frameworks for evaluating LLMs have been introduced, such as \texttt{lm-evaluation-harness} \citep{gao-2024-eval-harness}, OpenCompass \citep{contributors-2023-opencompass}, LightEval \citep{habib-2023-lighteval}, Inspect \citep{aisi-2024-inspect-ai} and \texttt{Unitxt} \citep{bandel-2024-unitxt}. However, while these tools provide useful infrastructure for running standardised benchmarks or implementing specific evaluation workflows, they typically do not aid users in selecting appropriate methods or in quantitatively measuring the effectiveness of these methods for a specific domain and task through meta-evaluation.

\footnotetext{Icons by Noun Project, authors Srinivas Agra, Iconiqu, Gonza Monta, Keyy Creative and suhaiba, \href{https://creativecommons.org/licenses/by/3.0/deed.en}{CC BY 3.0}.}

\begin{figure*}[tb]
    \centering
    \begin{subfigure}[m]{0.48\linewidth}
       \centering
        \includegraphics[width=0.99\textwidth]{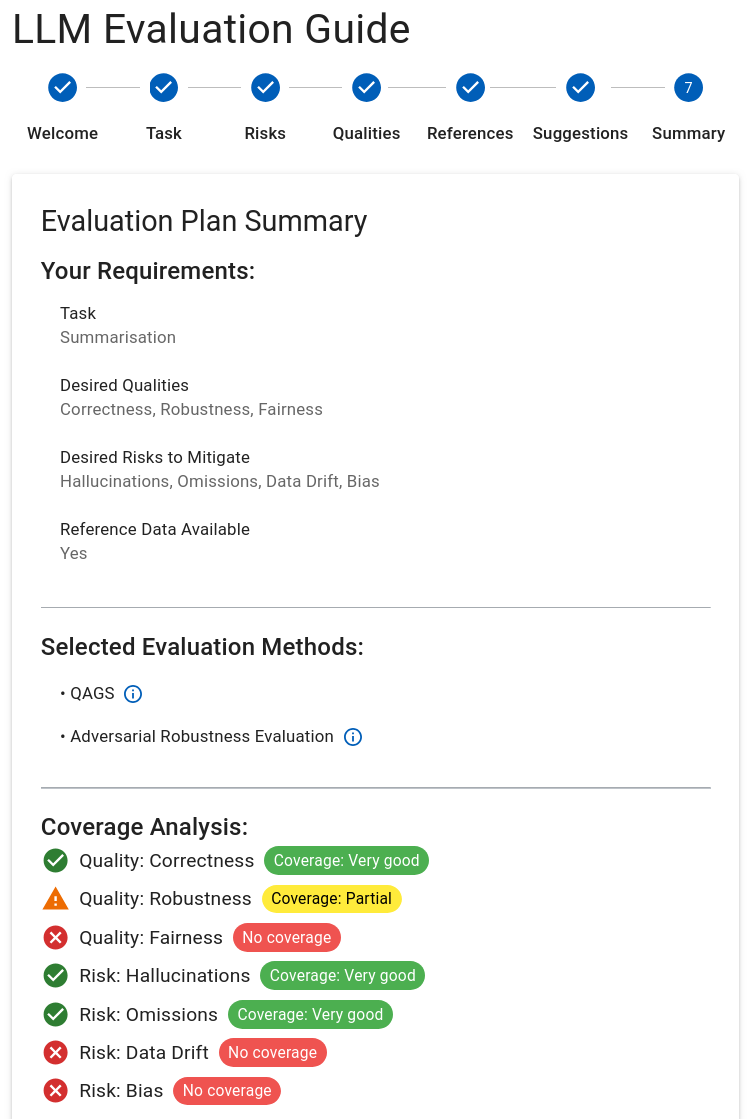}
        \caption{LLM Evaluation Guide}
        \label{fig:guide}
    \end{subfigure}
    \begin{subfigure}[m]{0.48\linewidth}
        \includegraphics[width=0.99\textwidth]{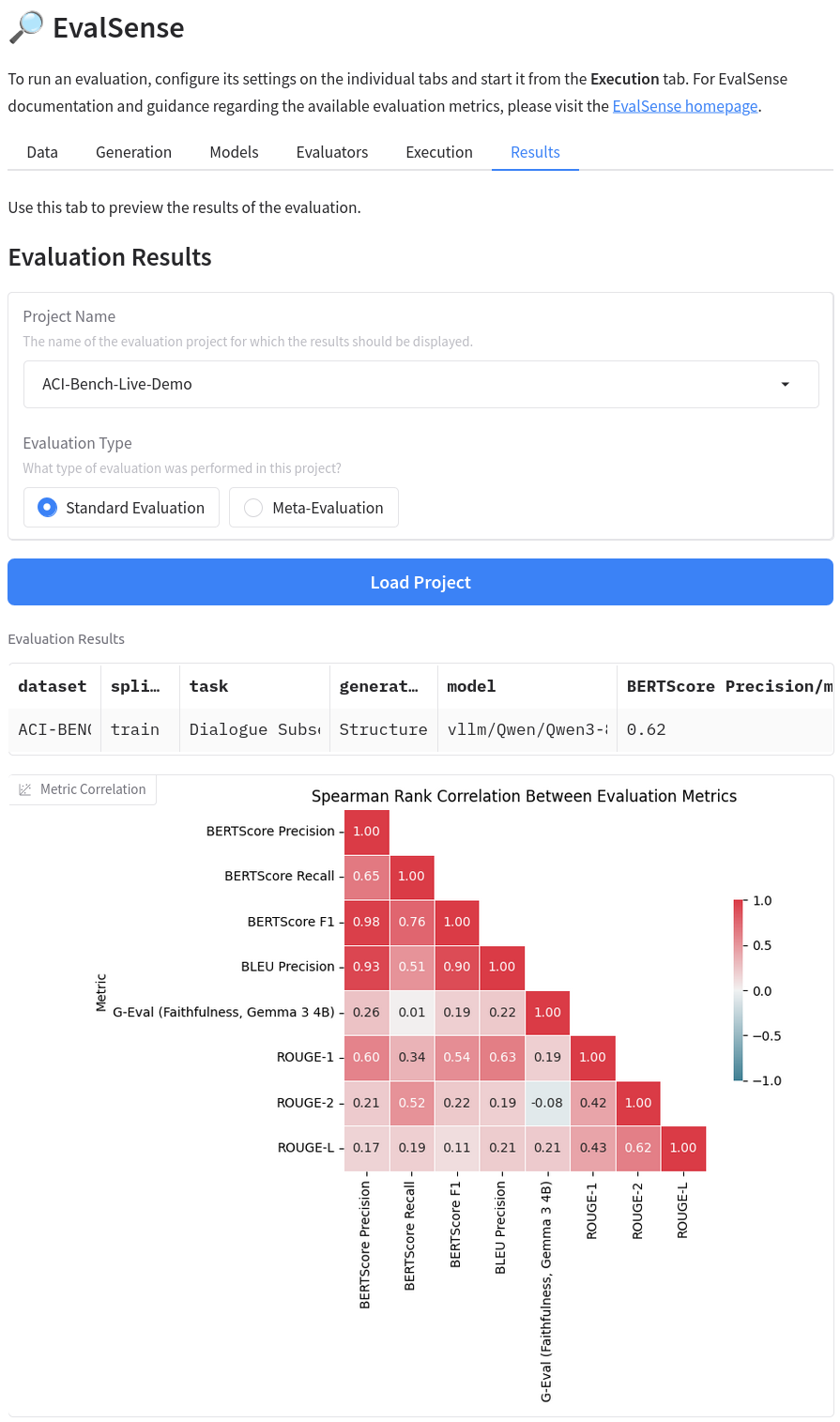}
        \caption{EvalSense user interface}
        \label{fig:ui}
    \end{subfigure}
    \caption{\textbf{(a)} EvalSense's LLM Evaluation Guide assists users in selecting suitable evaluation methods based on task-specific risks and requirements. The final evaluation plan summary highlights any risks and requirements not fully covered by the selected methods. The guide is available at \url{https://nhsengland.github.io/evalsense/guide}. \textbf{(b)} The web-based user interface provided by the EvalSense library can be used to configure and execute evaluations, as well as to view their results. Alternatively, this can be done through code after importing the library.}
    \label{fig:screenshots}
\end{figure*}

In response to these gaps, we introduce EvalSense, a highly flexible software framework that enables users to systematically evaluate LLMs on custom datasets. EvalSense offers two key features to help users navigate the spectrum of available evaluation methods:
\begin{enumerate}
    \item It includes an \emph{interactive evaluation guide}\footnote{Available on the EvalSense website at \url{https://nhsengland.github.io/evalsense/}.}, which prompts users to specify their tasks along with the associated risks and requirements, and then suggests appropriate evaluation strategies. After a subset of methods is selected, the guide generates a coverage report indicating whether the chosen methods comprehensively cover the specified risks and requirements (Figure \ref{fig:guide}).
    \item EvalSense incorporates \emph{automated meta-evaluation tools} that leverage controlled perturbations to validate evaluator reliability on the user's own dataset. These tools systematically degrade specific aspects of the output texts, verifying the degree to which these changes are reflected in the scores produced by the different evaluation techniques.
\end{enumerate}
In addition to these features, EvalSense also supports systematic experimentation, a broad range of local and API model providers, configuring evaluations through a graphical user interface (Figure \ref{fig:ui}), high-level result analysis and complex generation workflows.

To demonstrate and assess the capabilities of EvalSense, we apply it to a realistic evaluation task using ACI-Bench \citep{acibench-yim}, which involves generation of structured clinical notes from doctor-patient dialogues. Using EvalSense’s meta-evaluation tools, we demonstrate a non-trivial disparity in the quality of the scores produced by different evaluation methods. This demonstrates the importance of careful method selection and configuration, a process our framework is specifically designed to support.

Overall, we hope that EvalSense contributes to advancing best practices in LLM evaluation by systemizing the process of choosing between different evaluation strategies, both through the interactive guidance provided by the EvalSense guide and the quantitative meta-evaluation supported by the associated open-source library.

\section{Background and Related Work}

\subsection{Evaluation Metrics}
The growing use of machine learning models for text generation has led to the development of a wide range of evaluation techniques. Broadly, these can be categorised into three groups: traditional statistical metrics, LLM-as-a-judge methods and hybrid approaches.

\noindent \textbf{Statistical metrics} rely on direct, deterministic comparison of text units extracted from the evaluated text to the ground-truth reference. While still in use, these approaches are often overly simple to reliably assess the quality of open-ended texts. Examples of such metrics include the BLEU \citep{papineni-bleu} and ROUGE \citep{lin-rouge} scores.

\noindent \textbf{LLM-as-a-judge} methods leverage the general capabilities of LLMs to assess generated texts, mitigating many of the drawbacks associated with statistical metrics \cite{lianmin-2024-judging-judges}. However, the effectiveness of these methods is highly sensitive to the choice of model, prompt formulation, and general evaluation protocol. Notable examples include G-Eval \citep{liu-2023-g-eval} and GPTScore \citep{fu-2024-gptscore}.

\noindent \textbf{Hybrid methods} also make use of pre-trained models, but only use these models for targeted subtasks as part of a structured evaluation framework. For instance, BERTScore \citep{zhang-bertscore} compares texts based on contextual embeddings, while QAGS \citep{wang-2020-qags} assesses factual consistency using question answering.

\subsection{LLM Evaluation Toolkits}
A number of open-source toolkits have been introduced to support the evaluation of LLMs. Frameworks such as such as \texttt{lm-evaluation-harness} \citep{gao-2024-eval-harness}, OpenCompass \citep{contributors-2023-opencompass}, and LightEval \citep{habib-2023-lighteval} primarily focus on benchmarking models against standardised tasks and datasets. While some of these tools also support evaluation on custom data, they generally lack dedicated mechanisms for guiding evaluation design or assessing the suitability of selected methods in specific domains. The FreeEval framework \citep{yu-etal-2024-freeeval} extends beyond benchmarking by incorporating human judgements, bias detection, contamination analysis, and case-by-case inspection. However, it does not support automated meta-evaluation or provide interactive tools for evaluation strategy selection.

Among existing tools, Inspect \citep{aisi-2024-inspect-ai} is especially relevant to our work. EvalSense uses Inspect as its basis, inheriting its support for multiple model providers, tool use, agentic workflows and detailed logging infrastructure. Nevertheless, EvalSense significantly expands on this foundation through a more versatile and extensible pipeline tailored to custom datasets, improved resource management, support for advanced evaluation methods (including sophisticated LLM-as-a-judge and hybrid approaches), and its unique focus on meta-evaluation and domain-specific guidance. The bespoke components of EvalSense are described in the following section.

\section{EvalSense Pipeline}
\label{sec:pipeline}
EvalSense implements a robust and customisable pipeline that manages the key steps of the evaluation process, from data management and preprocessing, through LLM generation, to final evaluation and result analysis (as illustrated in Figure \ref{fig:pipeline}). Uniquely, the generation and result analysis modules provide built-in support for meta-evaluating the reliability of the used metrics in addition to simply returning their scores. The overall design of the pipeline supports reusability and extensibility by making individual components easily replaceable, enabling the use of custom datasets, LLMs, or evaluation methods.

\subsection{Dataset Manager}
Dataset managers are responsible for loading and generic preprocessing of the data on which the LLM is to be evaluated. For open-source datasets, this may involve downloading the data files from a publicly available repositories, while for internal datasets, the data will typically be loaded from a local file system or secure cloud storage. To simplify data management for custom datasets, EvalSense provides a base \href{https://nhsengland.github.io/evalsense/docs/api-reference/datasets/base/#evalsense.datasets.DatasetManager}{\texttt{DatasetManager}} class that defines a general interface for data managers and implements helper methods for retrieving associated files based on paths specified in a dataset configuration file. These methods can be overridden to support more complex data loading and preprocessing workflows.

\subsection{Task Preprocessor}
Task preprocessors implement any additional preprocessing steps that may be required to prepare the data for a specific task, as a single dataset may potentially support multiple such tasks. In many simple cases where no additional preprocessing is needed, users can rely on the \href{https://nhsengland.github.io/evalsense/docs/api-reference/tasks/base/#evalsense.tasks.DefaultTaskPreprocessor}{\texttt{DefaultTaskPreprocessor}}, which acts as an identify function. For more complex scenarios, users can define a custom preprocessing function following the \href{https://nhsengland.github.io/evalsense/docs/api-reference/tasks/base/#evalsense.tasks.TaskPreprocessingFunction}{\texttt{TaskPreprocessingFunction}} protocol, which can then be used with the standard \href{https://nhsengland.github.io/evalsense/docs/api-reference/tasks/base/#evalsense.tasks.TaskPreprocessor}{\texttt{TaskPreprocessor}}.

\subsection{LLM Generation Steps}
After preparing the data for the task, the pipeline generates LLM outputs for evaluation using pre-defined generation steps. These typically involve prompting the model with specific system and user prompts. Optionally, the generation steps can incorporate more advanced strategies, such as enabling access to external tools (e.g., via the Model Context Protocol\footnote{\url{https://modelcontextprotocol.io/introduction}}), incorporating model self-critiques \citep{madaan-2023-self-refine}, or using agentic workflows like ReAct \citep{yao-2023-react}. For the purposes of meta-evaluation, the LLM generation steps can also apply targetted perturbations degrading the quality of the output texts in predictable ways. In EvalSense, generation steps are defined via the \href{https://nhsengland.github.io/evalsense/docs/api-reference/generation/base/#evalsense.generation.GenerationSteps}{\texttt{GenerationSteps}} class, and the model configuration is specified using the \href{https://nhsengland.github.io/evalsense/docs/api-reference/generation/base/#evalsense.generation.ModelConfig}{\texttt{ModelConfig}}.

\subsection{Evaluator}
Evaluators implement automated methods for scoring model outputs based on predefined quality criteria. EvalSense includes several out-of-the-box score calculators, including:

\begin{itemize}
    \item BLEU (\href{https://nhsengland.github.io/evalsense/docs/api-reference/evaluation/evaluators/#evalsense.evaluation.evaluators.BleuPrecisionScoreCalculator}{\texttt{BleuPrecisionScoreCalculator}})
    \item ROUGE (\href{https://nhsengland.github.io/evalsense/docs/api-reference/evaluation/evaluators/#evalsense.evaluation.evaluators.RougeScoreCalculator}{\texttt{RougeScoreCalculator}})
    \item BERTScore (\href{https://nhsengland.github.io/evalsense/docs/api-reference/evaluation/evaluators/#evalsense.evaluation.evaluators.BertScoreCalculator}{\texttt{BertScoreCalculator}})
    \item G-Eval (\href{https://nhsengland.github.io/evalsense/docs/api-reference/evaluation/evaluators/#evalsense.evaluation.evaluators.GEvalScoreCalculator}{\texttt{GEvalScoreCalculator}})
    \item QAGS (\href{https://nhsengland.github.io/evalsense/docs/api-reference/evaluation/evaluators/#evalsense.evaluation.evaluators.QagsScoreCalculator}{\texttt{QagsScoreCalculator}})
\end{itemize}

These calculators can either be used independently or wrapped in an \href{https://nhsengland.github.io/evalsense/docs/api-reference/evaluation/base/#evalsense.evaluation.Evaluator}{\texttt{Evaluator}} class to be used as part of an evaluation pipeline. For convenience, EvalSense provides helper functions to easily initialise these evaluators (e.g., \href{https://nhsengland.github.io/evalsense/docs/api-reference/evaluation/evaluators/#evalsense.evaluation.evaluators.get_bleu_evaluator}{\texttt{get\_bleu\_evaluator}} for BLEU). All key aspects of the evaluator configurations, such as the used prompts and models for LLM-as-a-judge approaches are fully customisable. Users may also implement new evaluators to be used as part of the pipeline.

\subsection{Result Analyser}
While evaluators produce fine-grained scores for the individual samples and summary metrics for each configuration evaluated by the pipeline, result analysers can be used to summarise the results from multiple such configurations and surface higher-level trends. EvalSense currently includes three main analysers: \href{https://nhsengland.github.io/evalsense/docs/api-reference/workflow/analysers/#evalsense.workflow.analysers.TabularResultAnalyser}{\texttt{TabularResultAnalyser}} (for tabular summaries), \href{https://nhsengland.github.io/evalsense/docs/api-reference/workflow/analysers/#evalsense.workflow.analysers.MetricCorrelationAnalyser}{\texttt{MetricCorrelationAnalyser}} (for inter-metric correlation analysis), and \href{https://nhsengland.github.io/evalsense/docs/api-reference/workflow/analysers/#evalsense.workflow.analysers.MetaResultAnalyser}{\texttt{MetaResultAnalyser}}. The last-mentioned analyser is crucial for the meta-evaluation capabilities of EvalSense, and can assess the consistency of metric scores either with different levels of automated perturbations (increasingly degrading a specific aspect of the evaluated texts to obtain the ground-truth output rankings) or human annotations. Meanwhile, the correlation analysis provided by the \texttt{MetricCorrelationAnalyser} can be particularly helpful for identifying similarities between different evaluation methods.

\subsection{Pipeline}
All components are integrated through the \href{https://nhsengland.github.io/evalsense/docs/api-reference/workflow/base/#evalsense.workflow.Pipeline}{\texttt{Pipeline}}, which schedules and executes the planned experiments (i.e., different configurations to be evaluated). These experiments can be declared individually or in batches using the \href{https://nhsengland.github.io/evalsense/docs/api-reference/evaluation/base/#evalsense.evaluation.ExperimentConfig}{\texttt{ExperimentConfig}} and \href{https://nhsengland.github.io/evalsense/docs/api-reference/evaluation/base/#evalsense.evaluation.ExperimentBatchConfig}{\texttt{ExperimentBatchConfig}} data classes, enabling systematic evaluation sweeps. By default, the pipeline attempts to schedule the experiments in an optimal order to minimise the number of necessary model loads for local models, while also enabling users to resume any failed model generation tasks. As outlined above, the pipeline also supports automated meta-evaluation, performing controlled perturbations during the generation stage and assessing the reliability of different evaluation metrics during result analysis.

\subsection{Project}
All outputs, results, and metadata from pipeline execution are tracked through the \href{https://nhsengland.github.io/evalsense/docs/api-reference/workflow/base/#evalsense.workflow.Project}{\texttt{Project}} class, which maintains a record of all experiments associated with a given project, their status, and links to the relevant logs. This class also provides a high-level interface through which the pipeline and result analysers can access and update these logs.

\begin{table*}[tb]
    \centering
    \caption{Results from LLM evaluation on the ACI-Bench case study task using statistical and hybrid evaluation methods. Best results are bolded, second-best results are underlined.}
    \footnotesize
    \begin{adjustbox}{max width=\textwidth}
    \begin{tabular}{lccccccc}
        \toprule
        \textbf{Model} & \textbf{BLEU} & \textbf{ROUGE-1} & \textbf{ROUGE-2} & \textbf{ROUGE-L} & \textbf{BERTScore F1} & \textbf{Ternary QAGS} & \textbf{Judge QAGS} \\
        \midrule
        Gemma 3 12B & \textbf{0.128} & \underline{0.514} & \underline{0.207} & \textbf{0.300} & 0.666 & \underline{0.842} & \underline{0.817} \\
        Gemma 3 27B & 0.120 & 0.502 & 0.198 & 0.291 & \underline{0.668} & \textbf{0.846} & \textbf{0.821} \\
        Llama 3.1 8B & \underline{0.127} & \textbf{0.534} & \textbf{0.221} & \underline{0.294} & 0.662 & 0.806 & 0.789 \\
        Phi 4 14B & 0.120 & 0.504 & \underline{0.207} & 0.290 & \textbf{0.670} & 0.832 & 0.811 \\
        Qwen3 8B & 0.091 & 0.468 & 0.174 & 0.259 & 0.648 & 0.818 & 0.784 \\
        Qwen3 14B & 0.100 & 0.451 & 0.170 & 0.259 & 0.640 & 0.810 & 0.793 \\
        \bottomrule
    \end{tabular}
    \end{adjustbox}
    \label{tab:case-study-traditional}
\end{table*}

\begin{table*}[tb]
    \centering
    \caption{Results from LLM evaluation on the ACI-Bench case study task using different variants of G-Eval. Best results are bolded, second-best results are underlined}
    \footnotesize
    \begin{adjustbox}{max width=\textwidth}
	\begin{tabular}{lcccccc}
		\toprule
		\textbf{Model} & \textbf{Brief Gemma 3} & \textbf{Det. Gemma 3} & \textbf{Brief Llama 3.1} & \textbf{Det. Llama 3.1} & \textbf{Brief Qwen3} & \textbf{Det. Qwen3} \\
		\midrule
        Gemma 3 12B & \textbf{0.929} & \underline{0.904} & \underline{0.847} & 0.834 & 0.777 & \underline{0.665} \\
        Gemma 3 27B & \underline{0.926} & \textbf{0.916} & \textbf{0.848} & 0.840 & \underline{0.798} & 0.640 \\
        Llama 3.1 8B & 0.835 & 0.823 & 0.788 & 0.801 & 0.683 & 0.598 \\
        Phi 4 14B & 0.906 & 0.892 & 0.826 & 0.836 & 0.763 & 0.662 \\
        Qwen3 8B & 0.864 & 0.876 & 0.845 & \textbf{0.854} & 0.775 & 0.630 \\
        Qwen3 14B & 0.885 & 0.899 & 0.843 & \underline{0.849} & \textbf{0.814} & \textbf{0.682} \\
		\bottomrule
    \end{tabular}
    \end{adjustbox}
    \label{tab:case-study-g-eval}
\end{table*}

\begin{table}[tb]
    \centering
    \caption{Results from perturbation-based meta-evaluation of the different evaluation methods. Methods are ordered from best to worst.}
    \footnotesize
    \begin{tabular}{l c}
        \toprule
        \textbf{Method Name} & \textbf{Avg. Correlation} \\
        \midrule
        G-Eval (Detailed, Gemma 3 27B) & 0.999 \\
        G-Eval (Brief, Gemma 3 27B) & 0.998 \\
        G-Eval (Detailed, Llama 3.1 8B) & 0.995 \\
        G-Eval (Brief, Llama 3.1 8B) & 0.992 \\
        Ternary QAGS (Llama 3.1 8B) & 0.982 \\
        Judge QAGS (Llama 3.1 8B) & 0.969 \\
        G-Eval (Brief, Qwen 3 14B) & 0.967 \\
        G-Eval (Detailed, Qwen 3 14B) & 0.924 \\
        BERTScore F1 & 0.431 \\
        ROUGE-1 & 0.323 \\
        BLEU Precision & 0.296 \\
        ROUGE-L & 0.232 \\
        ROUGE-2 & 0.049 \\
        \bottomrule
    \end{tabular}
    \label{tab:case-study-meta}
\end{table}

\section{Evaluation Case Study}
\label{sec:evaluation-case-study}
\paragraph{Task Setup} To demonstrate EvalSense's effectiveness, we apply it to LLM evaluation on the task of dialogue summarisation using the ACI-Bench dataset \citep{acibench-yim}. In this setting, the LLM is tasked with generating a structured clinical note based on a doctor-patient dialogue transcript. Given that correctness and comprehensiveness of the generated notes are the most crucial qualities in this context, our evaluation primarily focuses on these aspects.

We used the $120$ samples from the test partitions of the ACI-Bench dataset. Since we are using the original, unchanged dialogues from the dataset, the dataset manager and task preprocessor stages of our pipeline are mostly focused on loading the relevant samples without significant additional preprocessing.

For the LLM generation steps, we used the system and user prompts from \cite{kanithi-medic}. The user prompt specified the intended note structure and section headings, as more general instructions would make the output format ambiguous. We experimented with six different open-weight models: Llama 3.1 8B \citep{dubey-2024-llama-3}, Phi 4 \citep{abdin-2024-phi-4}, Qwen3 8B and Qwen3 14B \citep{yang-2024-qwen-3}, and Gemma 3 12B and Gemma 3 27B \citep{kamath-2025-gemma-3}. All models were run locally using vLLM \citep{kwon-2023-vllm} in their default precisions, with the generation temperature set to $0.7$, top-p sampling value of $0.95$ and a seed of $42$.

\paragraph{Evaluation Setup} The case study involved a total of $13$ variants of five major evaluators implemented in EvalSense: BLEU \citep{papineni-bleu}, ROUGE \citep{lin-rouge}, BERTScore \citep{zhang-bertscore}, G-Eval \citep{liu-2023-g-eval} and QAGS \cite{wang-2020-qags}. For ROUGE, we considered ROUGE-1, ROUGE-2 and ROUGE-L. The G-Eval metric was used with two different prompt variations: a detailed prompt providing thorough instructions on how to evaluate a note and a brief prompt asking the model to evaluate general faithfulness and accuracy. We also experimented with different G-Eval judge models: Llama 3.1 8B, Qwen3 14B and Gemma 3 27B. For the QAGS metric, we considered two different versions: ternary QAGS that generates questions requiring ternary responses and judge QAGS using more open-ended questions with an LLM judge comparing the responses. Both considered variants of the QAGS score used Llama 3.1 8B as the underlying model.

For our meta-evaluation, we used a set of three prompts instructing the model to apply different levels of perturbations to the note: one rephrasing the note without changing its meaning, one introducing minor content changes and one significantly changing the meaning of the note. The used prompts are given in Appendix \ref{apd:prompts}.

\paragraph{Results} The results of our evaluation are summarised in Tables~\ref{tab:case-study-traditional} and~\ref{tab:case-study-g-eval}. We can observe that there is substantial disagreement among the different methods, with no universally best-performing model. Without further information on each method's reliability for this task, drawing definitive conclusions would be difficult.

However, based on the meta-evaluation results in Table~\ref{tab:case-study-meta}, we can assign greater weight to G-Eval variants using Gemma 3 and Llama 3.1, as well as both QAGS versions. These methods consistently rank the Gemma 3 models highest, except for G-eval with Llama 3.1 using the detailed prompt. Notably, statistical metrics and BERTScore underperform compared to LLM-based methods.

\section{Conclusion}
In this paper, we introduced EvalSense, a novel framework for systematic evaluation of LLMs on custom tasks. Unlike other toolkits, which mainly focus on direct application of evaluation methods without providing principled ways to assess their suitability, EvalSense guides users in selecting evaluation approaches tailored to their specific domains and provides quantitative insights about the effectiveness of these approaches through meta-evaluation. We demonstrated its capabilities through a case study on structured clinical note generation from doctor-patient dialogues, showing that it supports robust evaluation even when different evaluation methods yield disagreeing results.

\section*{Acknowledgments}
We thank the UK AI Security Institute and the wider development team for their work on the Inspect framework, which serves as a basis for EvalSense.

\bibliography{custom}

\appendix

\section{Used prompts}
\label{apd:prompts}

\subsection{Note Generation Prompt}
The prompt used for the ACI-Bench note genration, adapted from \citep{kanithi-medic}, is shown in Listing \ref{lst:note-generation}.

\subsection{Perturbation Prompt 1}
The prompt used for rephrasing the output notes without changing their meaning is shown in Listing \ref{lst:perturbation-1}.

\subsection{Perturbation Prompt 2}
The prompt used for introducing minor content changes is shown in Listing \ref{lst:perturbation-2}.

\subsection{Perturbation Prompt 3}
The prompt used for significantly changing the meaning of the generated notes is given in Listing \ref{lst:perturbation-3}.

\begin{lstfloat*}
\caption{Note generation prompt}
\begin{lstlisting}
Your task is to generate a clinical note based on a conversation between a doctor and a patient. Use the following format for the clinical note:

1. **CHIEF COMPLAINT**: [Brief description of the main reason for the visit]
2. **HISTORY OF PRESENT ILLNESS**: [Summary of the patient's current health status and any changes since the last visit]
3. **REVIEW OF SYSTEMS**: [List of symptoms reported by the patient]
4. **PHYSICAL EXAMINATION**: [Findings from the physical examination]
5. **RESULTS**: [Relevant test results]
6. **ASSESSMENT AND PLAN**: [Doctor's assessment and plan for treatment or further testing]

**Conversation:**
{prompt}

**Note:**
\end{lstlisting}
\label{lst:note-generation}
\end{lstfloat*}

\begin{lstfloat*}
\caption{Perturbation prompt 1}
\begin{lstlisting}
Your task is to generate a clinically plausible variation of the provided clinical note. 

You should maintain the original note's structure and formatting, but modify its content according to the specified types of perturbation below. Try to maintain internal consistency and general medical plausibility when applying any changes.

**Perturbation Instructions**  
Apply the following types of perturbations:
- Rephrase sentences while preserving the exact medical meaning. You may use synonyms, vary sentence structure, or change sentence length, but all clinical facts and measurements must remain unchanged.
- Slightly alter the writing style, such as using different terminology or presenting findings differently, while ensuring the factual content remains identical.

Respond only with the perturbed clinical note, do not include any commentary, reasoning or explanation.

**Original Clinical Note**
{prompt}

**Perturbed Clinical Note**
\end{lstlisting}
\label{lst:perturbation-1}
\end{lstfloat*}

\begin{lstfloat*}
\caption{Perturbation prompt 2}
\begin{lstlisting}
Your task is to generate a clinically plausible variation of the provided clinical note. 

You should maintain the original note's structure and formatting, but modify its content according to the specified types of perturbation below. Try to maintain internal consistency and general medical plausibility when applying any changes.

**Perturbation Instructions**  
Apply the following types of perturbations:
- Make small changes to test results and quantitative measurements, ensuring they remain clinically plausible and consistent with the original context.
- Introduce minor modifications to the patient's reported symptoms, making sure they are still consistent with the assessment, diagnosis, and treatment plan (e.g., adding or substituting symptoms that commonly co-occur).
- Slightly adjust the patient's clinical history, ensuring consistency with the assessment, diagnosis, and treatment plan.
- Make minor modifications to the treatment plan, but ensure it remains appropriate for the assessment and diagnosis.

Respond only with the perturbed clinical note, do not include any commentary, reasoning or explanation.

**Original Clinical Note**
{prompt}

**Perturbed Clinical Note**
\end{lstlisting}
\label{lst:perturbation-2}
\end{lstfloat*}

\begin{lstfloat*}
\caption{Perturbation prompt 3}
\begin{lstlisting}
Your task is to generate a clinically plausible variation of the provided clinical note. 

You should maintain the original note's structure and formatting, but modify its content according to the specified types of perturbation below. Try to maintain internal consistency and general medical plausibility when applying any changes.

**Perturbation Instructions**  
Apply the following types of perturbations:
- Significantly alter test results and quantitative measurements, in a way that may change the clinical interpretation or implications of the note.
- Make substantial changes to the patient's reported symptoms, potentially affecting the clinical interpretation of the note.
- Make substantial changes to the patient's clinical history, potentially affecting the clinical interpretation.
- Significantly modify the treatment plan, such that it may lead to a different clinical outcome than the original plan.

Respond only with the perturbed clinical note, do not include any commentary, reasoning or explanation.

**Original Clinical Note**
{prompt}

**Perturbed Clinical Note**
\end{lstlisting}
\label{lst:perturbation-3}
\end{lstfloat*}

\end{document}